\documentclass[final]{cvpr}

\usepackage{times}
\usepackage{epsfig}

\usepackage{graphicx}

\usepackage{amsmath}
\usepackage{amssymb}

\usepackage[caption = false]{subfig}
\usepackage{comment}
\usepackage{color}

\usepackage{dsfont}
\usepackage{multirow}
\usepackage{cite}

\usepackage{verbatim}
\usepackage{pseudocode}

\usepackage{array}
\usepackage{booktabs, caption, makecell}

\usepackage{algorithm}
\usepackage{algpseudocode}%

\usepackage[pagebackref=true,breaklinks=true,colorlinks,bookmarks=false]{hyperref}

\newcolumntype{L}[1]{>{\raggedright\arraybackslash}m{#1}}
\newcolumntype{C}[1]{>{\centering\arraybackslash}m{#1}}
\newcolumntype{R}[1]{>{\raggedleft\arraybackslash}m{#1}}
\newcolumntype{+}{>{\global\let\currentrowstyle\relax}}
\newcolumntype{^}{>{\currentrowstyle}}

\newcommand{\RomNum}[1]{\MakeUppercase{\romannumeral #1}}

\def\cvprPaperID{4585} 
\def\confYear{CVPR 2021}
\begin{document}

\title{Harmonious Semantic Line Detection via Maximal Weight Clique Selection}

\author{Dongkwon Jin, Wonhui Park\\
Korea University\\
{\tt\footnotesize dongkwonjin, whpark@mcl.korea.ac.kr}
\and
Seong-Gyun Jeong\\
42dot.ai\\
{\tt\footnotesize seonggyun.jeong@42dot.ai}
\and
Chang-Su Kim\\
Korea University\\
{\tt\footnotesize changsukim@korea.ac.kr}
}

\maketitle

\begin{abstract}
   A novel algorithm to detect an optimal set of semantic lines is proposed in this work. We develop two networks: selection network (S-Net) and harmonization network (H-Net). First, S-Net computes the probabilities and offsets of line candidates. Second, we filter out irrelevant lines through a selection-and-removal process. Third, we construct a complete graph, whose edge weights are computed by H-Net. Finally, we determine a maximal weight clique representing an optimal set of semantic lines. Moreover, to assess the overall harmony of detected lines, we propose a novel metric, called HIoU. Experimental results demonstrate that the proposed algorithm can detect harmonious semantic lines effectively and efficiently. Our codes are available at \href{https://github.com/dongkwonjin/Semantic-Line-MWCS}{https://github.com/dongkwonjin/Semantic-Line-MWCS}.
\end{abstract}

\section{Introduction}
\begin{figure}[t]

    \subfloat {\includegraphics[width=1.66cm,height=1.3cm]{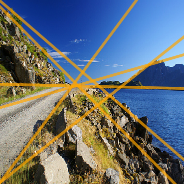}}\,\!\!
    \subfloat {\includegraphics[width=1.66cm,height=1.3cm]{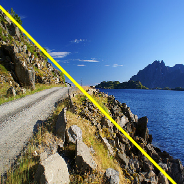}}\,\!\!
    \subfloat {\includegraphics[width=1.66cm,height=1.3cm]{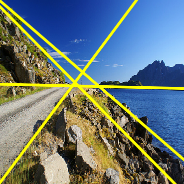}}\,\!\!
    \subfloat {\includegraphics[width=1.66cm,height=1.3cm]{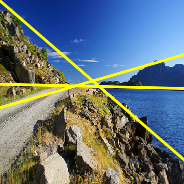}}\,\!\!
    \subfloat {\includegraphics[width=1.66cm,height=1.3cm]{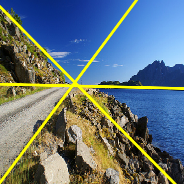}}\\[-4.9ex]

    \subfloat {\includegraphics[width=1.66cm,height=1.3cm]{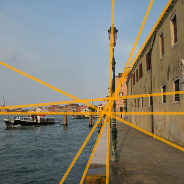}}\,\!\!
    \subfloat {\includegraphics[width=1.66cm,height=1.3cm]{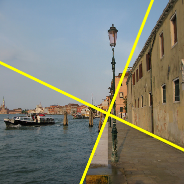}}\,\!\!
    \subfloat {\includegraphics[width=1.66cm,height=1.3cm]{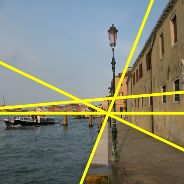}}\,\!\!
    \subfloat {\includegraphics[width=1.66cm,height=1.3cm]{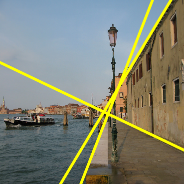}}\,\!\!
    \subfloat {\includegraphics[width=1.66cm,height=1.3cm]{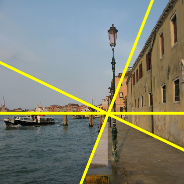}}\\[-4.9ex]

    \setcounter{subfigure}{0}
    \subfloat[] {\includegraphics[width=1.66cm,height=1.3cm]{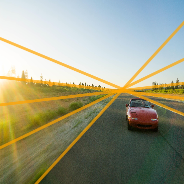}}\,\!\!
    \subfloat[] {\includegraphics[width=1.66cm,height=1.3cm]{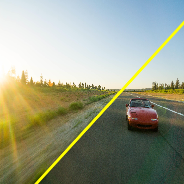}}\,\!\!
    \subfloat[] {\includegraphics[width=1.66cm,height=1.3cm]{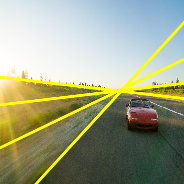}}\,\!\!
    \subfloat[] {\includegraphics[width=1.66cm,height=1.3cm]{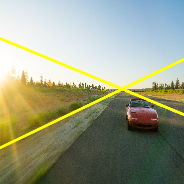}}\,\!\!
    \subfloat[] {\includegraphics[width=1.66cm,height=1.3cm]{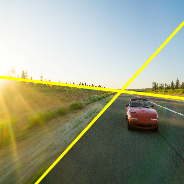}}
    \caption
    {
        In each scene, straight lines approximating region boundaries are shown in (a). Among them, three subsets of lines are shown in (b), (c), and (d), which are insufficient, over-segmenting, and sub-optimal for describing the composition of the scene, respectively. In contrast, an optimal set of semantic lines in (e) convey the composition of the scene harmoniously.
    }
    \label{fig:semantic_lines}
\end{figure}

A \textit{semantic line}~\cite{lee2017,jin2020} is defined as a meaningful line, separating different semantic regions in a scene, which is approximated by an end-to-end straight line. A group of semantic lines in an image can be regarded as optimal, when they convey the composition of the image harmoniously, as shown in Figure~\ref{fig:semantic_lines}(e). Thus, in an optimal set, the lines should harmonize with one another.

Semantic lines provide important visual cues in high-level image understanding \cite{Freeman2007,krages2012,lee2018photographic,guo2012,hillel2014,lee2017_vpgnet,zhou2019_nips}.
In photography, semantic lines, such as horizontal, vertical, and symmetric ones, are essential composition components. Harmony of such lines are closely related to subjective  quality of a photograph \cite{Freeman2007,krages2012,lee2018photographic}. In autonomous driving systems \cite{guo2012,hillel2014,hou2020_inter}, boundaries of road lanes and sidewalks should be detected reliably to control vehicle maneuvers, which can be also described by semantic lines. Moreover, dominant parallel lines intersect at vanishing points~\cite{lee2017_vpgnet,zhou2019_nips} under perspective projection, conveying depth impression. They are also semantic lines~\cite{jin2020}. However, it is challenging to detect semantic lines, which are often unobvious and implied by complex boundaries of semantic regions.

\begin{figure*}[t]
  \centering
  \includegraphics[width=1\linewidth]{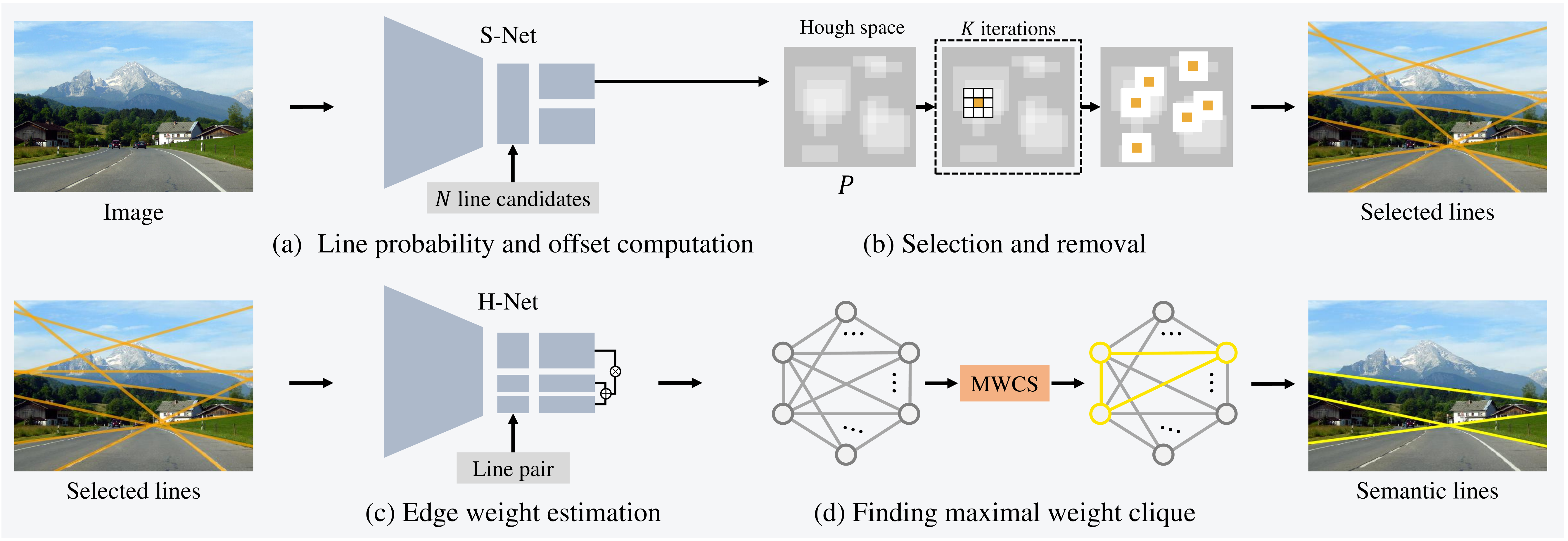}
  \caption{Overview of the proposed algorithm.}

  \label{fig:Overview_fig}
\end{figure*}

Many techniques have been developed to detect line segments in a scene by exploiting hand-crafted features \cite{matas2000,von2008,desolneux200,akinlar2011} or deep features \cite{huang2018,xue2019,zhou2019_line,lin2020}. However, they may extract redundant short line segments or focus on identifying obvious line structures in man-made environments. Recently, several attempts have been made to detect semantic lines~\cite{workman2016,zhai2016, lee2017, jin2020, han2020}. Horizon lines, which are a specific type of semantic lines, have been estimated by CNN-based methods~\cite{workman2016,zhai2016}. In~\cite{lee2017, jin2020, han2020}, semantic line detectors have been proposed. They have two stages: line detection and refinement. In the detection stage, deep line features are extracted to classify line candidates, but implied lines may be undetected or the computational cost for extracting discriminative features can be too high. In the refinement stage, redundant lines are removed through non-maximum suppression (NMS) or pairwise comparison. Although these techniques provide promising results, they may fail to consider the harmony between detected lines and thus may yield sub-optimal results, as shown in Figure~\ref{fig:semantic_lines}(d).

In this paper, a novel algorithm to detect an optimal set of harmonious semantic lines is proposed based on maximal weight clique selection (MWCS). We formulate the detection as finding a maximal weight clique in a complete graph~\cite{Graph1998,chartrand2019}. To this end, we design two networks: selection network (S-Net) and harmonization network (H-Net). Given an image and a set of line candidates, S-Net first computes the classification probability and regression offsets of each candidate. Second, we filter out irrelevant lines by performing a selection-and-removal process. Third, we construct a complete graph, in which the node set contains the selected lines. H-Net computes its edge weights. Finally, we determine a maximal weight clique representing harmonious semantic lines. Experimental results demonstrate that the proposed algorithm can detect harmonious semantic lines accurately and efficiently.

This work has the following major contributions:
\begin{itemize}
\itemsep0mm
\item We formulate the semantic line detection as finding an maximal weight clique in a complete graph.
\item We develop two networks, S-Net and H-Net, to construct the complete graph.
\item We introduce a novel metric, called HIoU, to assess the overall harmony of semantic lines, which is more reasonable than the existing metrics in \cite{lee2017,han2020}.
\item The proposed algorithm yields competitive semantic line detection performance to the state-of-the-art DRM technique~\cite{jin2020}, while reducing the computational complexity by a factor of $\frac{1}{20}$.
\end{itemize}

\section{Related Work}

\subsection{Line segment detection}
Line segments give important visual cues for image semantics. In line segment detection \cite{matas2000,von2008,desolneux200,akinlar2011}, many short segments are detected using low-level features, such as image gradients. This approach, however, may not discriminate meaningful lines from noisy ones. To utilize higher-level features, deep learning methods have been proposed \cite{huang2018,xue2019,zhou2019_line,lin2020}.
In \cite{huang2018}, a line heat map and junctions were predicted by networks. Then, a wireframe was obtained by connecting the junctions based on the heat map. In \cite{zhou2019_line}, a line candidate was generated by connecting two junctions and then was classified into either a salient one or not. In \cite{xue2019}, attraction field maps were computed by a network to deal with local ambiguity and class imbalance in line segment detection. In \cite{lin2020}, a network was trained with a Hough transform block to combine local information with global line priors. These methods \cite{huang2018,xue2019,zhou2019_line,sun2019} focus on detecting obvious lines in man-made environments.

\subsection{Semantic line detection}
Semantic lines, located near the boundaries of semantic regions, represent the layout and composition of images. Several methods \cite{workman2016,zhai2016,diaz2019,lee2017,han2020,jin2020} have been developed to detect implied but semantically meaningful lines. In~\cite{workman2016,zhai2016,diaz2019}, horizon lines were detected by CNNs, which were refined by exploiting vanishing points or using soft labels of line parameters. In~\cite{lee2017}, Lee \etal proposed the first semantic line detector. They devised a line pooling layer to extract local features along each line candidate. Those features were fed into classification and regression layers to detect semantic lines. Then, an NMS scheme was performed to remove redundant lines, based on the edge detector \cite{xie_2015_ICCV}. In~\cite{jin2020}, Jin~\etal extracted more discriminative line features by designing a region pooling layer and the mirror attention module. Then, they selected the most semantic lines and removed redundant lines alternately through pairwise ranking and matching. In~\cite{han2020}, Han~\etal transformed line features into a Hough parametric space to facilitate parallel processing of multiple line candidates. Then, they trained a network to predict a line probability map, which was used to determine semantic lines by computing the centroids of connected components.

\subsection{Road lane detection}
In autonomous driving systems, it is important to reliably detect the boundaries of road lanes, sidewalks, or crosswalks. Early methods \cite{he2004,aly2008,hillel2014,zhou2010} used hand-crafted low-level features to extract lanes. Recently, to cope with complicated road scenes, attempts have been made to detect road lanes using deep semantic segmentation frameworks \cite{pan2018,hou2019_road,hou2020_inter,qin2020}. In~\cite{pan2018}, Pan~\etal proposed a network to learn spatial relationship of lanes through message passing between convolution layers. In~\cite{hou2019_road}, a network was designed to generate attention maps at different layers, which were used to refine the output of deeper ones. In~\cite{hou2020_inter}, the inter-region affinity graph was constructed to transfer  structural relationship between lanes from teacher to student networks. In~\cite{qin2020}, to achieve a high processing speed, a network was developed to identify the location of each lane on a predefined set of rows only.

\section{Proposed Algorithm}

Figure~\ref{fig:Overview_fig} is an overview of the proposed algorithm, which contains S-Net and H-Net. First, given an image and a set of line candidates, S-Net computes the line probability and the regression offsets of each candidate. Second, irrelevant candidates are filtered out through a selection-and-removal process. Third, a complete graph, whose node set consists of the selected lines, is constructed and its edge weights are computed by H-Net. Finally, a maximal weight clique, representing harmonious semantic lines, is determined.

\subsection{Problem formulation}
\begin{figure}[t]

  \centering
  \includegraphics[width=1\linewidth]{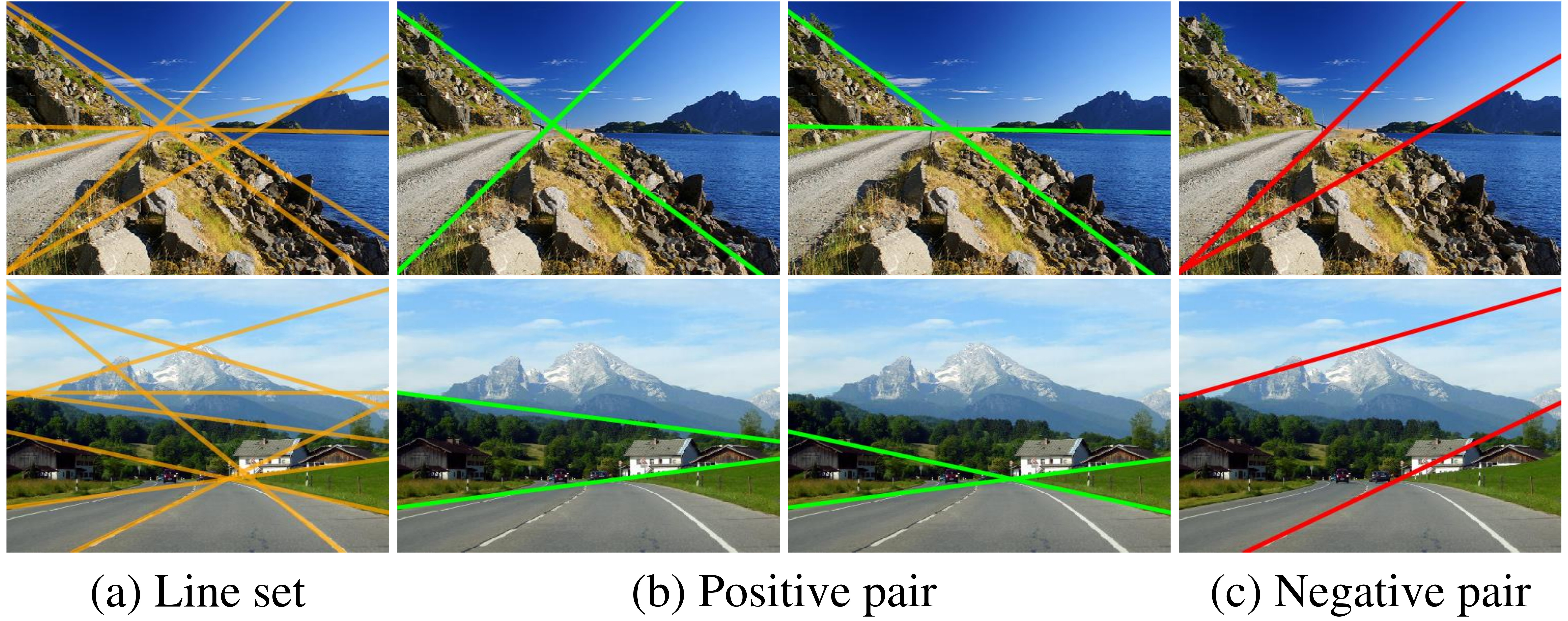}
  \caption{In a line set, a pair of lines can be harmonious and draw visual attention to meaningful regions as in (b). In contrast, they can be redundant or inharmonious as in (c).}
  \label{fig:line_pair}
\end{figure}

Semantic lines in an image can be regarded as optimal if they convey the composition of the image harmoniously. In other words, in an optimal set, every pair of semantic lines should harmonize with each other. As in Figure~\ref{fig:line_pair}(b), a pair of semantic lines should direct visual attention to meaningful regions. In contrast, in Figure~\ref{fig:line_pair}(c), two lines are redundant or inharmonious.
Based on this observation, we formulate the semantic line detection as finding a maximal weight clique in a complete graph~\cite{Graph1998,chartrand2019}. In the complete graph, detected lines form the node set, and each edge weight represents how harmonious the associated two lines are. Thus, by finding a maximal weight clique, we find an optimal set of harmonious semantic lines.

\subsection{Node selection: filtering line candidates}
\label{ssec:s-net}

It is computationally infeasible to construct a complete graph for all line candidates. Therefore, we select reliable nodes only by filtering line candidates.

\vspace{0.15cm}
\noindent\textbf{Line candidate generation:}
A line candidate, which is an end-to-end straight line in an image, can be parameterized by polar coordinates in the Hough space~\cite{kiryati1991,han2020,lin2020}.
Let $\mathbf{l}=(\rho, \varphi)$ denote a line, where $\rho$ is its distance from the center of the image and $\varphi$ is its angle from the $x$-axis. Then, we generate $N$ line candidates, denoted by $\mathbf{l}_{n}=(\rho_n, \varphi_n)$, $1\leq n \leq N$, by quantizing $\rho$ and $\varphi$ uniformly.

\begin{figure}[t]
  \centering
  \includegraphics[width=1\linewidth]{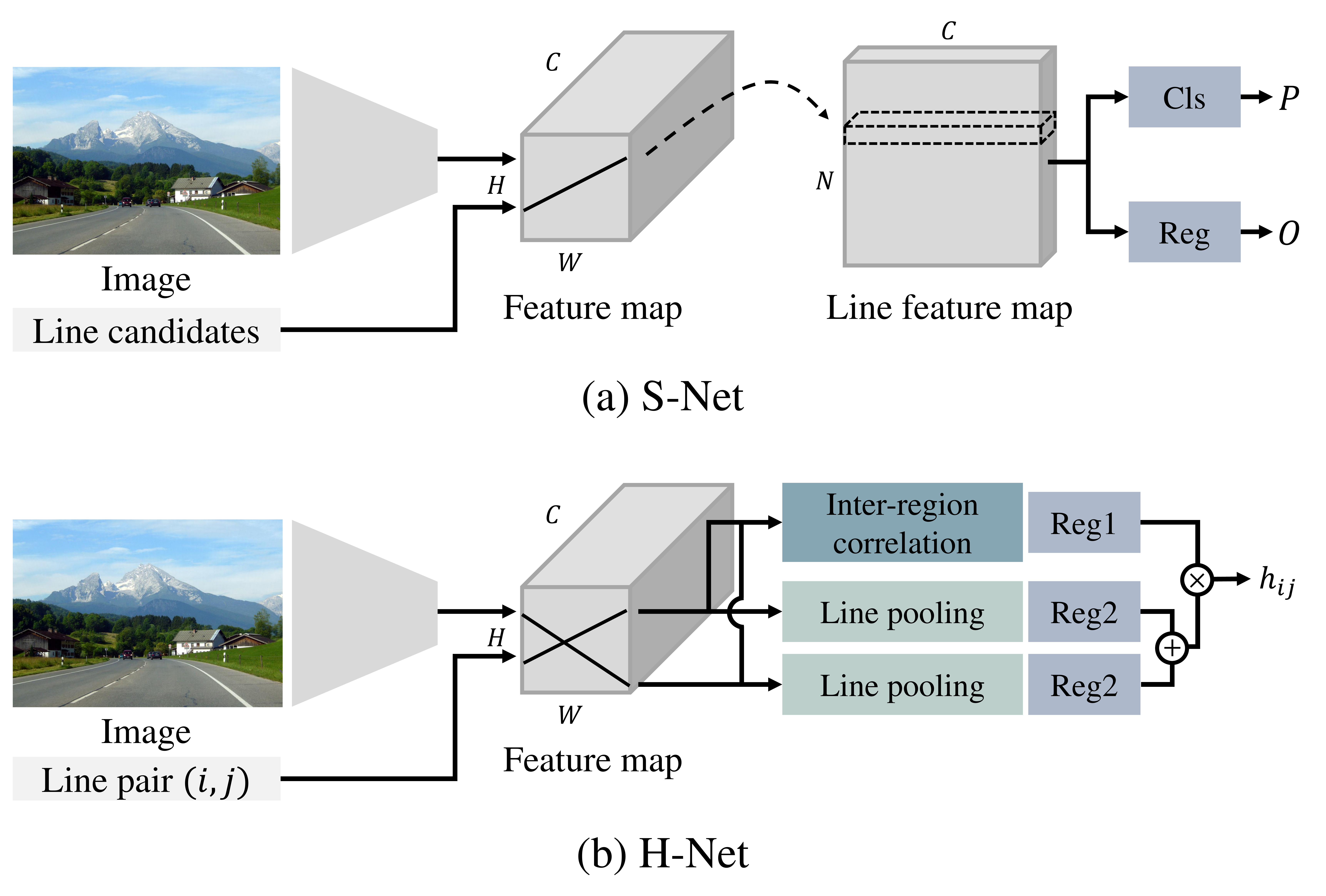}
  \caption{Architecture of S-Net and H-Net: (a) S-Net takes an image and line candidates and extracts the line feature map. The classification probabilities and regression offsets of the line candidates are then computed by two fully connected layers. (b) H-Net takes an image and a line pair $(i, j)$ to extract local and inter-region features. Two types of regression layers are used to compute the harmony score~$h_{ij}$.}
  \label{fig:Network_fig}
\end{figure}

\vspace{0.15cm}
\noindent\textbf{S-Net:}
For each line candidate, we compute its classification probability and regression offsets. To this end, we develop S-Net based on the conventional line detectors~\cite{lee2017, han2020, jin2020}. Figure~\ref{fig:Network_fig}(a) shows the architecture of S-Net. From an image, S-Net extracts a convolutional feature map $X=[X^1, X^2, \ldots, X^C]\in \mathbb{R}^{H\times W\times C}$, where $H$, $W$, and $C$ denote the feature height, the feature width, and the number of channels. Then, the line feature map $Y=[Y^1, Y^2, \ldots, Y^C]\in \mathbb{R}^{N\times C}$ is obtained by averaging the features of pixels along $\mathbf{l}_n$;
\begin{equation}\label{eq:line_pooling}
    \textstyle
    {Y^c({n})} = \frac{1}{|\mathbf{l}_n|} \sum_{\mathbf{p} \in {\mathbf{l}_n}}{X^c}(\mathbf{p})
\end{equation}
for $1\leq n \leq N$ and $1 \leq c \leq C$, where $|\mathbf{l}_n|$ denotes the number of pixels along $\mathbf{l}_n$. We then obtain the probability vector $P$ and the line offset matrix $O$ by
\begin{equation}
    \textstyle
    P = \sigma(f_1(Y)) \quad \mbox{and} \quad
    O = f_2(Y)
\end{equation}
where $f_1$ and $f_2$ are fully-connected layers of sizes $C\times 1$ and $C\times 2$ for classification and regression, respectively, and $\sigma(\cdot)$ is the sigmoid activation function. For the $n$th line candidate $\mathbf{l}_{n}=(\rho_n, \varphi_n)$, $P_{n}$ indicates the probability that it is semantic, and $O_{n}= \Delta \mathbf{l}_n = ({\Delta \rho_n, \Delta \varphi_n})$ is the offset vector for line refinement in Section~\ref{ssec:GCS}.

The architecture and training process of S-Net are described in detail in the supplemental document.

\begin{figure*}[t]
\vspace{-0.4cm}
    \subfloat[] {\includegraphics[width=2.47cm,height=1.7cm]{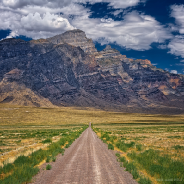}}\,\!\!
    \subfloat[] {\includegraphics[width=2.47cm,height=1.7cm]{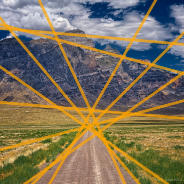}}\,\!\!
    \subfloat[] {\includegraphics[width=2.47cm,height=1.7cm]{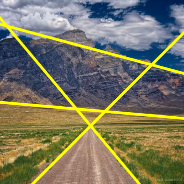}}\,\!\!
    \subfloat[] {\includegraphics[width=2.47cm,height=1.7cm]{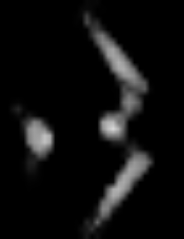}}\,\!\!
    \subfloat[] {\includegraphics[width=2.47cm,height=1.7cm]{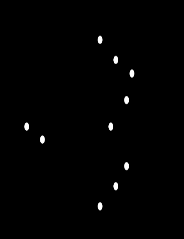}}\,\!\!
    \subfloat[] {\includegraphics[width=2.47cm,height=1.7cm]{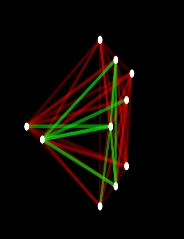}}\,\!\!
    \subfloat[] {\includegraphics[width=2.47cm,height=1.7cm]{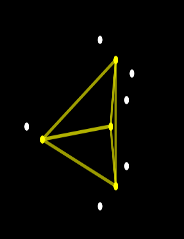}}
    \caption
    {
        Illustration of the proposed algorithm: (a) input image, (b) selected lines through the selection-and-removal process, (c) semantic lines, (d) probabilities of line candidates in the Hough space, (e) node set corresponding to the selected lines, (f) complete graph, (g) maximal clique. In (f), an edge is depicted in green or red depending on whether its weight is above the threshold $\kappa$ in \eqref{eq:tau} or not. As the weight approaches zero, the transparency increases.
    }
    \label{fig:overall_vis}
\end{figure*}

\vspace*{0.15cm}
\noindent\textbf{Selection and removal:} In the conventional algorithms \cite{lee2017, han2020, jin2020}, to detect semantic lines, only the line candidates with probabilities higher than a threshold are selected and then post-processed (\eg non-maximum suppression). However, this may cause false negatives, which have low probabilities because of being implicit but are semantic nonetheless. To reduce such false negatives, instead of thresholding, we perform the selection-and-removal process in Figure~\ref{fig:Overview_fig}(b). We select the most reliable line $\mathbf{l}_{i^\star}$ by
\begin{equation}
    \textstyle
    i^\star =  \arg \max_{i} P_i
\end{equation}
and then remove overlapping lines with the selected one. Specifically, we remove 24 lines within the $5\times 5$ grid centered at $\mathbf{l}_{i^\star}$ in the Hough space~\cite{han2020,lin2020}. We perform this process $K$ times to compose the node set of $K$ selected lines. Figure~\ref{fig:overall_vis}(b) and (e) show such selected lines on the image and Hough spaces, respectively.

\subsection{Edge weighting: harmony score estimation}
\label{ssec:h-net}

\noindent\textbf{Inter-region correlation:} To tell positive pairs in Figure~\ref{fig:line_pair}(b) from negative pairs in Figure~\ref{fig:line_pair}(c), we design the inter-region-correlation (IRC) module that analyzes the regions separated by a pair of lines.

Let ${\cal R}_i$, $1\leq i \leq M$, denote the regions separated by two lines. There can be three or four regions, \ie~$M=3$ or $4$. We extract the regional feature vector $\mathbf{r}_i$ of ${\cal R}_i$ by
\begin{equation}
    \textstyle
    \mathbf{r}_i = \frac{1}{|{\cal R}_i|} \sum_{\mathbf{p} \in {{\cal R}_i}} X(\mathbf{p}).
\end{equation}
We compute the softmax probability $a_i$ of the area $|{\cal R}_i|$ to scale the regional feature vectors, and then concatenate the scaled vectors into
\begin{equation}
\textstyle
R=[a_1{\bf r}_1, a_2{\bf r}_2, a_3{\bf r}_3, a_4{\bf r}_4]
\end{equation}
of size $C\times 4$. If $M=3$, we fill in the rightmost vector with zeros.
Then, $R$ is fed into a fully connected layer to yield the IRC feature.

\vspace*{0.15cm}
\noindent\textbf{H-Net:} We develop H-Net using the IRC module. It takes an image and a pair of lines, indexed by $i$ and $j$, to yield the harmony score $h_{ij}$ ranging from 0 to 1. Figure~\ref{fig:Network_fig}(b) shows the H-Net architecture. The convolution layers of VGG16~\cite{Simonyan2015} are used as the feature extractor, which is followed by three parallel branches of the IRC module and line pooling layers. We employ the line pooling layers to perform the pooling in~(\ref{eq:line_pooling}) for lines $i$ and $j$, respectively. We use two types of regression layers: one for yielding the IRC score  of the two lines (Reg1), and the other for computing unary reliability of each line (Reg2). Finally, we compute the harmony score $h_{i j}$ by multiplying the IRC score with the average of the unary reliability levels.

We configure the training data for H-Net as follows. It is assumed that every pair of ground-truth semantic lines in an image harmonize with each other. Thus, we declare such pairs as positive, while the others as negative. In other words, a line pair $(i,j)$ is positive only if both lines $i$ and $j$ are semantic. Then, the harmony score $\bar{h}_{ij}$ is annotated as 1 or 0 depending on whether the pair $(i,j)$ is positive or not. However, this strict definition of a positive pair causes a class imbalance: there are too few positive pairs. Thus, we disturb the line locations of each positive pair and annotate the corresponding harmony score $\bar{h}_{ij}$ to be proportional to $e^{-(d_i^2+d_j^2)}$, where $d_i$ and $d_j$ denote the disturbances of lines $i$ and $j$. Also, the loss function for training H-Net is defined as $\ell_{\rm H} = (h_{ij}-\bar{h}_{ij})^2$,
where ${\bar h}_{ij}$ is the ground-truth harmony score and $h_{ij}$ is its estimate. The supplemental document describes the training process and architecture of H-Net in more detail.

\subsection{Graph optimization: finding harmonious lines}
\label{ssec:GCS}

\noindent\textbf{Graph construction:} We construct a complete graph $G=({\cal V}, {\cal E})$, in which the node set ${\cal V}=\{v_1, v_2, \ldots, v_{K}\}$ represents the $K$ lines selected using S-Net in Section~\ref{ssec:s-net}. Every pair of lines are connected by an edge in the edge set ${\cal E}=\{(v_i, v_j): i \neq j\}$. Each edge is assigned a weight $w(v_i, v_j)=h_{ij}$ by H-Net in Section~\ref{ssec:h-net}. Figure~\ref{fig:overall_vis}(f) visualizes a complete weighted graph.

\vspace{0.15cm}
\noindent\textbf{MWCS:} As mentioned earlier, a set of semantic lines is optimal, if any two lines in the set are harmonious with each other. Thus, finding such an optimal set is equivalent to finding a clique of nodes~\cite{Graph1998}, which are mutually connected and have a maximal sum of weights (\ie harmony scores).

Let $\theta$ denote a clique, represented by the index set of member nodes. Then, we define the harmonization energy $E_{\rm harmony}(\theta)$ of clique $\theta$ as
\begin{equation}
    \textstyle
    E_{\rm harmony}(\theta) = \sum_{i \in \theta} \sum_{j \in \theta, j>i} w(v_i, v_j)
\end{equation}
which is the sum of all edge weights in $\theta$. Finding the clique that maximizes this energy is NP-hard~\cite{feremans2003generalized}. However, in this work, $K$ is set to be a small number. The default $K$ is 8. There are about $2^K$ possible cliques, which are also manageable. Thus, exhaustive search is adopted to find a maximal weight clique. First, we generate the set of possible cliques $\Theta = \{\theta_t\}$ in the graph $G$, where each clique $\theta_t$ consists of more than two nodes. Then, we select the maximal weight clique $\theta^{\star}$ that maximizes the harmonization energy:
\begin{equation}\label{eq:maximal_clique}
    \textstyle
    \theta^{\star} = \arg \max_{\theta_t \in \Theta} E_{\rm harmony}(\theta_t)
\end{equation}
subject to a constraint
\begin{equation}
\textstyle
{\min_{i,j\in \theta} w(v_i, v_j)}>\kappa
\label{eq:tau}
\end{equation}
where $\kappa$ is a threshold. If there is no clique satisfying the constraint, we select the maximal single-node clique $\theta^{\star} = \{ i^{\star} \}$ by
\begin{equation}\label{eq:reliability_node}
    i^{\star} =  \arg \max_{i} h_{ii}.
\end{equation}
The self-harmony score $h_{ii}$ is obtained by applying the same line as duplicated input to H-Net.

After obtaining the set of harmonious semantic lines, we refine each line by
\begin{equation}
\mathbf{l}_{v_{i}} + \Delta \mathbf{l}_{v_{i}}
\end{equation}
where $\Delta \mathbf{l}_{v_{i}}$ denotes the offset vector, generated by the regression layer of S-Net. Figure~\ref{fig:overall_vis}(c) and (g) show the set of harmonious semantic lines on the image and Hough spaces.

\begin{figure*}[t]

  \centering
  \includegraphics[width=1\linewidth]{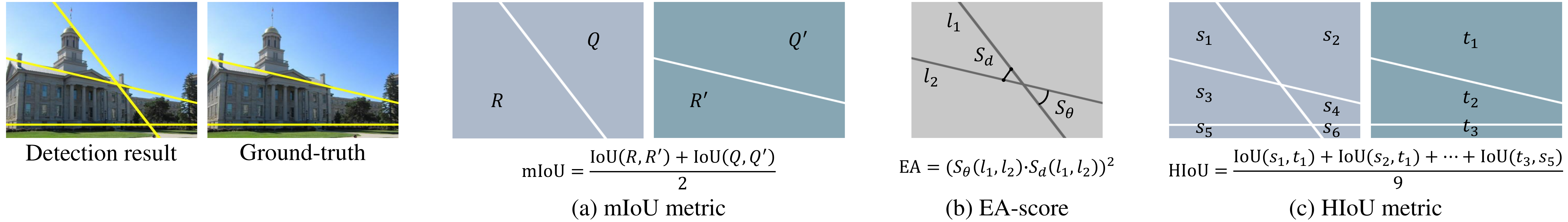}
  \\[-5pt]
  \caption{Illustration of two existing metrics of mIoU \cite{lee2017} and EA-score\cite{han2020} and the proposed HIoU metric.}
  \label{fig:metric_fig}
\end{figure*}

\begin{figure}[t]
  \centering
  \includegraphics[width=1\linewidth]{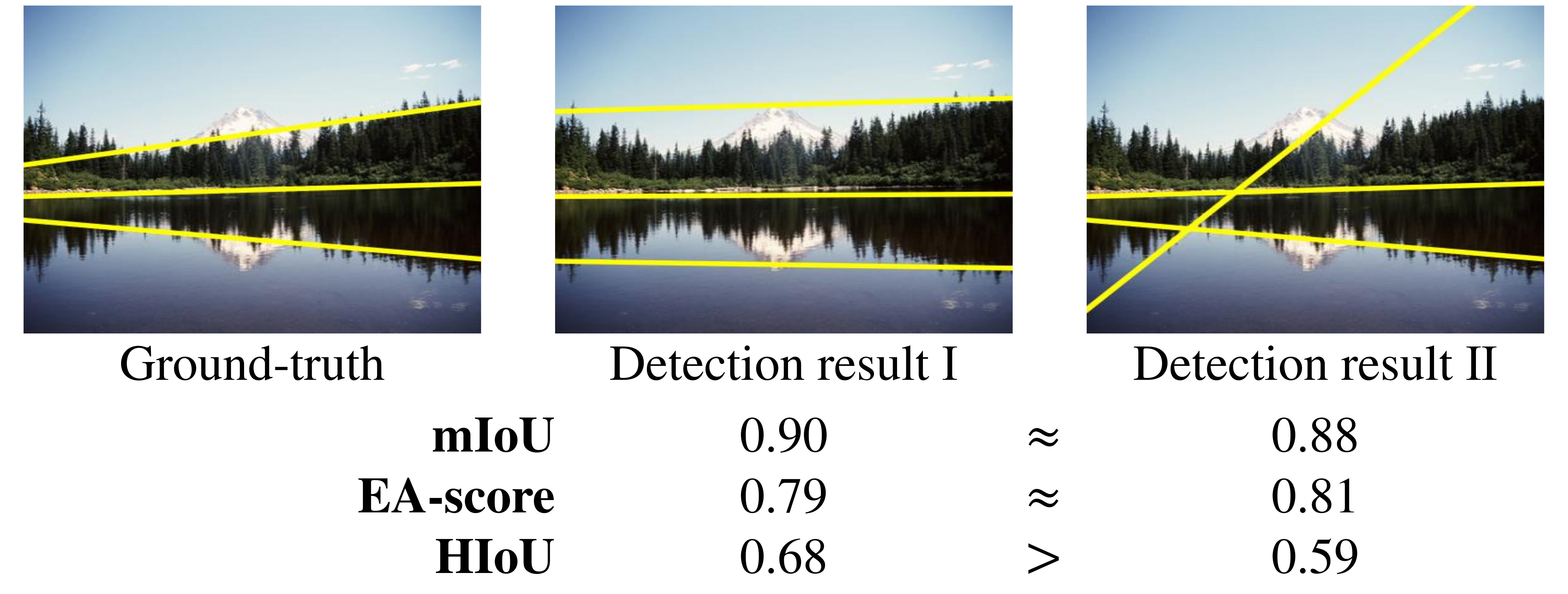}
  \\[-5pt]
  \caption{Comparison of mIoU \cite{lee2017}, EA-score \cite{han2020}, and the proposed HIoU metric: There are two detection results for the same ground-truth. In result \RomNum{1}, the position of each detected line is different from the ground-truth, but the detected lines convey the composition of the image relatively well. In result \RomNum{2}, two detected lines match the ground-truth exactly, but they are not harmonious with the remaining one. As a group, they are inferior to result \RomNum{1}. Since mIoU and EA-score consider only the accuracy of each individual line, they do not tell the difference between these two results and provide only marginally different scores. In contrast, HIoU quantifies the superiority of result \RomNum{1} correctly.}
  \label{fig:metric_comparison}
\end{figure}

\section{Experimental Results}

\subsection{Datasets}

\noindent\textbf{SEL~\cite{lee2017}:} It is the first semantic line dataset, containing 1,750 outdoor images, which are split into 1,575 training and 175 testing images. Each semantic line is annotated by the coordinates of two end points on an image boundary.

\vspace{0.15cm}
\noindent\textbf{SEL\_Hard~\cite{jin2020}:}
It is a more challenging dataset for testing semantic line detectors. It contains 300 test images, selected from the ADE20K segmentation dataset~\cite{zhou2017_ade}. Its  semantic lines are less obvious and more severely occluded in more cluttered scenes.

\vspace{0.15cm}
\noindent\textbf{SL5K~\cite{kai2020}:}
It is a rich and diverse dataset in terms of the number of lines and scene categories. It is composed of 4,000 training and 1,000 testing images.

\vspace{0.15cm}
\noindent\textbf{CULane~\cite{pan2018}:} It is a dataset for road lane detection, containing 88,000 training images. Its 34,680 test images are classified into 9 categories. For each image, the pixel-wise mask for up to 4 road lanes is provided. The proposed algorithm is tested on 3,911 test images in the `no lane' category, in which each lane is highly implied or even invisible.

\subsection{Metrics}

\noindent\textbf{Conventional metrics:}
There are two existing metrics to assess semantic line detection results: mIoU ~\cite{lee2017} and EA-score \cite{han2020}.
In the mIoU metric, a detected line is regarded as correct if its mIoU score with the ground-truth semantic line is greater than a threshold~$\tau$ as illustrated in Figure~\ref{fig:metric_fig}(a). In the EA-score, a detected line is regarded as correct if its similarity with the ground-truth is greater than the threshold as shown in Figure~\ref{fig:metric_fig}(b). The similarity is composed of two factors $S_d$ and $S_\theta$, which are based on the Euclidean distance between the midpoints of the lines and the angular distance of the lines, respectively. In both metrics, the precision and the recall are computed by
\begin{equation}\label{eq:pre_rec}
    \textstyle
    {\rm Precision} = \frac{N_l}{N_l + N_e}, \;\;\; {\rm Recall} = \frac{N_l}{N_l + N_m}
\end{equation}
where $N_l$ is the number of correctly detected semantic lines, $N_e$ is the number of false positives, and $N_m$ is the number of false negatives. Then, the F-measure is computed by
\begin{equation}\label{eq:fscore}
    \textstyle
    {\rm F \text{-}\rm measure} =  \frac{2 \times \rm Precision \times \rm Recall}{\rm Precision + \rm Recall}.
\end{equation}
The area under curve (AUC) performances of the precision, recall, F-measure curves are measured in the entire range of the threshold~$\tau$, which are denoted by AUC$\_$P, AUC$\_$R, and AUC$\_$F, respectively~\cite{lee2017}.

However, these metrics measure only the positional accuracy of each detected line. They do not consider how harmonious multiple detected lines are with one another in a scene. Hence, they may yield misleading scores, as exemplified in  Figure~\ref{fig:metric_comparison}.

\begin{table*}[t]\centering
    \renewcommand{\arraystretch}{0.85}
    \vspace*{0.1cm}
    \caption
    {
        Comparison of the AUC and HIoU scores (\%) on the SEL and SEL\_Hard datasets. The processing speeds in frames per second (fps) are also compared. For the AUC scores, the mIoU metric is used.
    }
    \vspace*{-0.15cm}
    \resizebox{0.92\linewidth}{!}{
    \begin{tabular}[t]{+L{2.2cm}^C{1.2cm}^C{1.2cm}^C{1.2cm}^C{1.2cm}^C{1.2cm}^C{1.2cm}^C{1.2cm}^C{1.2cm}^C{1.2cm}}
    \toprule
    \multirow{2}{*}{}  & \multicolumn{4}{c}{SEL} & \multicolumn{4}{c}{SEL\_Hard} & \multirow{2}{*}{fps}\\
    \cmidrule(lr){2-5} \cmidrule(lr){6-9}
    & AUC\_P & AUC\_R & AUC\_F & HIoU & AUC\_P & AUC\_R & AUC\_F & HIoU\\
    \midrule
         SLNet~\cite{lee2017}           & 80.72 & \underline{84.50} & 82.57 & 77.87 & 74.22 & 70.68 & 72.41 & 59.71 & 7.35\\
         DHT~\cite{han2020}             & \underline{89.27} & 78.53 & 83.56 & 79.62 & 83.55 & 67.98 & 75.09 & 63.39 & \bf{30.30}\\
         DRM~\cite{jin2020}             & 85.44 & \bf{87.16} & \underline{86.29} & \underline{80.23} & \underline{87.19} & \bf{77.69} & \bf{82.17} & \bf{68.83} & 1.05\\
         Proposed                       & \bf{89.61} & 84.21 & \bf{86.83} & \bf{81.03} & \bf{87.60} & \underline{72.56} & \underline{79.38} & \underline{65.99} & \underline{21.74}\\
    \bottomrule
    \end{tabular}}
    \label{table:auc_sel}
    \vspace*{-0.05cm}
\end{table*}

\vspace*{0.15cm}
\noindent\textbf{HIoU metric:}
We propose the harmony-based intersection-over-union (HIoU) metric to assess the overall harmony of detected lines. Detected lines tend to convey harmonious impression about the composition of an image, when their division of the image is consistent with the division by the ground-truth. Suppose that the set of detected lines and the set of ground-truth lines divide the image into regions $S=\{s_1, s_2, \ldots, s_{N}\}$ and $T=\{t_1, t_2, \ldots, t_{M}\}$, respectively. Then, we define HIoU as
\begin{equation}\label{eq:hiou}
    \textstyle
    {\rm HIoU} = \frac{\sum_{i=1}^{N} \max_{k}{{\rm IoU}(s_i, t_k)} + \sum_{j=1}^{M} \max_{k}{{\rm IoU}(t_j, s_k)}}{N+M}.
\end{equation}
In other words, for each $s_i$, we find the matching $t_k$ and measure their IoU. Similarly, for each $t_j$, we find its IoU with the matching $s_k$. Then, the average of these bi-directional matching IoU's becomes the HIoU score. Figure~\ref{fig:metric_fig}(c) illustrates how to compute an HIoU score. Figure~\ref{fig:metric_comparison} shows that HIoU assesses detected lines more reasonably than the existing metrics do, by considering the harmony among the detected lines.

\begin{figure}[t]

    \vspace{-0.40cm}
    \centering
    \setcounter{subfigure}{0}
    \subfloat {\includegraphics[width=2.9cm, height=2.9cm]{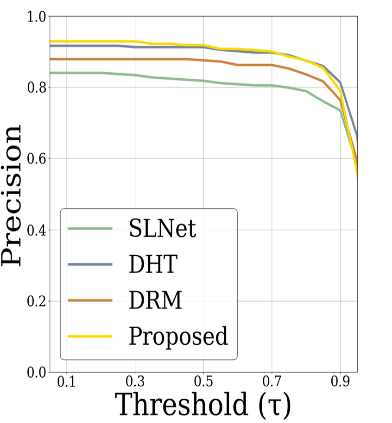}}
    \subfloat {\includegraphics[width=2.9cm, height=2.9cm]{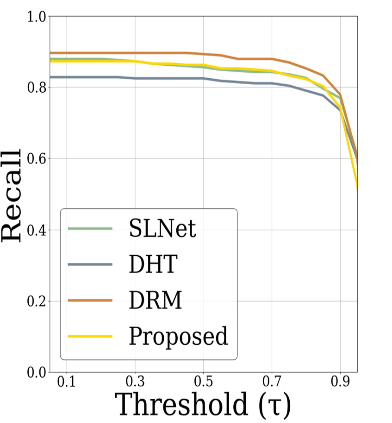}}
    \subfloat {\includegraphics[width=2.9cm, height=2.9cm]{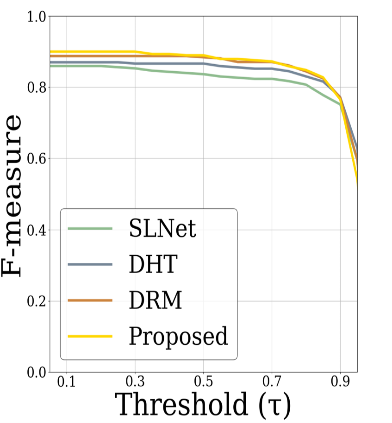}}
    \caption
    {
        Comparison of the precision, recall, and F-measure curves in terms of the threshold $\tau$ on the SEL dataset. The mIoU metric is used.
    }

    \label{fig:auc_sel}
\end{figure}

\captionsetup[subfigure]{labelformat=empty}
\begin{figure*}[t]
\vspace{-0.3cm}
\setlength{\belowcaptionskip}{-0.4cm}
    \begin{flushright}

    \subfloat {\raisebox{1.7em}{\rotatebox[origin=t]{90}{\scriptsize Ground-truth}}}\hspace{-0.01cm}\,
    \subfloat {\includegraphics[width=2.42cm,height=1.55cm]{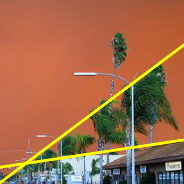}}\,\!\!
    \subfloat {\includegraphics[width=2.42cm,height=1.55cm]{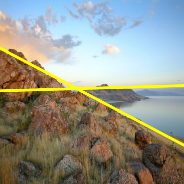}}\,\!\!
    \subfloat {\includegraphics[width=2.42cm,height=1.55cm]{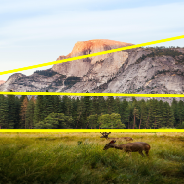}}\,\!\!
    \subfloat {\includegraphics[width=2.42cm,height=1.55cm]{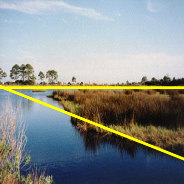}}\,\!\!
    \subfloat {\includegraphics[width=2.42cm,height=1.55cm]{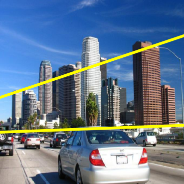}}\,\!\!
    \subfloat {\includegraphics[width=2.42cm,height=1.55cm]{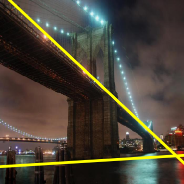}}\,\!\!
    \subfloat {\includegraphics[width=2.42cm,height=1.55cm]{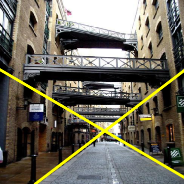}}\\[-2.2ex]

    \subfloat {\raisebox{1.7em}{\rotatebox[origin=t]{90}{\scriptsize Proposed}}}\,\!
    \subfloat {\includegraphics[width=2.42cm,height=1.55cm]{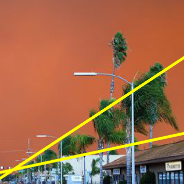}}\,\!\!
    \subfloat {\includegraphics[width=2.42cm,height=1.55cm]{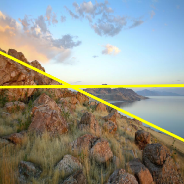}}\,\!\!
    \subfloat {\includegraphics[width=2.42cm,height=1.55cm]{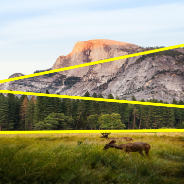}}\,\!\!
    \subfloat {\includegraphics[width=2.42cm,height=1.55cm]{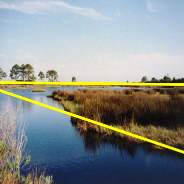}}\,\!\!
    \subfloat {\includegraphics[width=2.42cm,height=1.55cm]{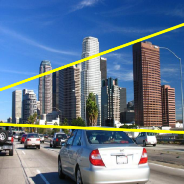}}\,\!\!
    \subfloat {\includegraphics[width=2.42cm,height=1.55cm]{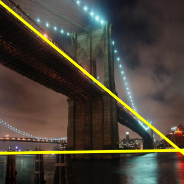}}\,\!\!
    \subfloat {\includegraphics[width=2.42cm,height=1.55cm]{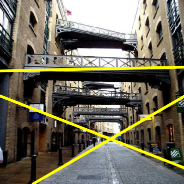}}\\[-2.2ex]

    \setcounter{subfigure}{0}
    \subfloat {\raisebox{1.7em}{\rotatebox[origin=t]{90}{\scriptsize DRM~\cite{jin2020}}}}\,\!
    \subfloat {\includegraphics[width=2.42cm,height=1.55cm]{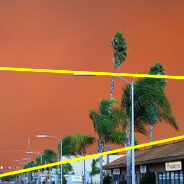}}\,\!\!
    \subfloat {\includegraphics[width=2.42cm,height=1.55cm]{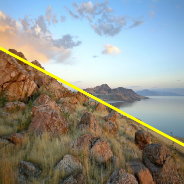}}\,\!\!
    \subfloat {\includegraphics[width=2.42cm,height=1.55cm]{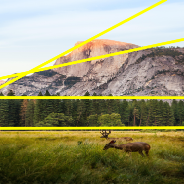}}\,\!\!
    \subfloat {\includegraphics[width=2.42cm,height=1.55cm]{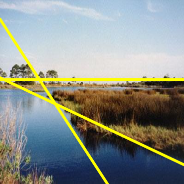}}\,\!\!
    \subfloat {\includegraphics[width=2.42cm,height=1.55cm]{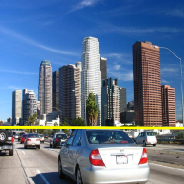}}\,\!\!
    \subfloat {\includegraphics[width=2.42cm,height=1.55cm]{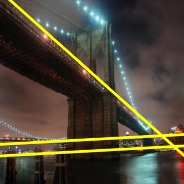}}\,\!\!
    \subfloat {\includegraphics[width=2.42cm,height=1.55cm]{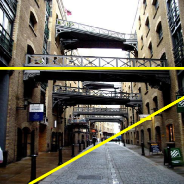}}\\[-2.2ex]

    \subfloat {\raisebox{1.7em}{\rotatebox[origin=t]{90}{\scriptsize DHT~\cite{han2020}}}}\,\!
    \subfloat {\includegraphics[width=2.42cm,height=1.55cm]{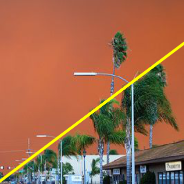}}\,\!\!
    \subfloat {\includegraphics[width=2.42cm,height=1.55cm]{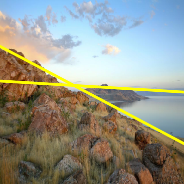}}\,\!\!
    \subfloat {\includegraphics[width=2.42cm,height=1.55cm]{figure/result/sel_result/1713_DHT.png}}\,\!\!
    \subfloat {\includegraphics[width=2.42cm,height=1.55cm]{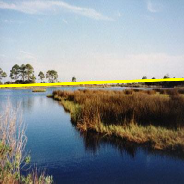}}\,\!\!
    \subfloat {\includegraphics[width=2.42cm,height=1.55cm]{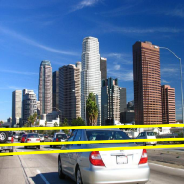}}\,\!\!
    \subfloat {\includegraphics[width=2.42cm,height=1.55cm]{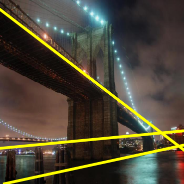}}\,\!\!
    \subfloat {\includegraphics[width=2.42cm,height=1.55cm]{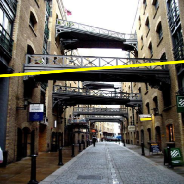}}\\[-2.2ex]

    \subfloat {\raisebox{1.7em}{\rotatebox[origin=t]{90}{\scriptsize SLNet~\cite{lee2017}}}}\,\!
    \subfloat {\includegraphics[width=2.42cm,height=1.55cm]{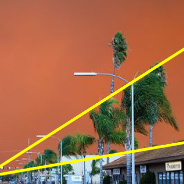}}\,\!\!
    \subfloat {\includegraphics[width=2.42cm,height=1.55cm]{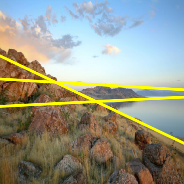}}\,\!\!
    \subfloat {\includegraphics[width=2.42cm,height=1.55cm]{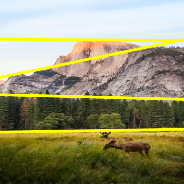}}\,\!\!
    \subfloat {\includegraphics[width=2.42cm,height=1.55cm]{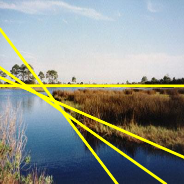}}\,\!\!
    \subfloat {\includegraphics[width=2.42cm,height=1.55cm]{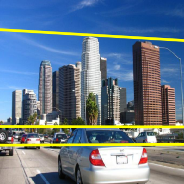}}\,\!\!
    \subfloat {\includegraphics[width=2.42cm,height=1.55cm]{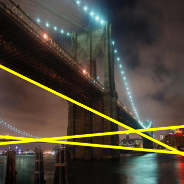}}\,\!\!
    \subfloat {\includegraphics[width=2.42cm,height=1.55cm]{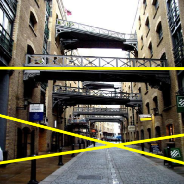}}\\

    \caption
    {
        Comparison of semantic line detection results. The left three images are from SEL, and the others from SEL\_Hard.
    }
    \label{fig:sel_result}
    \end{flushright}
\end{figure*}
\captionsetup[subfigure]{labelformat=parens}

\subsection{Comparative assessment}
We compare semantic line detection results of the proposed algorithm with those of the conventional SLNet~\cite{lee2017}, DHT~\cite{han2020}, and DRM~\cite{jin2020}.

\vspace*{0.15cm}
\noindent\textbf{Comparison on SEL:}
Figure~\ref{fig:auc_sel} compares the precision, recall, and F-measure curves of the proposed algorithm and the conventional algorithms on the SEL dataset. Table~\ref{table:auc_sel} reports the AUC performances of these curves. The proposed algorithm provides a poorer recall but a better precision than the conventional algorithms. F-measure is the harmonic mean of recall and precision. Note that the proposed algorithm outperforms all conventional algorithms in terms of F-measure and HIoU.

\vspace*{0.15cm}
\noindent\textbf{Comparison on SEL\_Hard:}
Table~\ref{table:auc_sel} also compares the results on SEL\_Hard. For this comparison as well, we use the same algorithms that are trained using the training images in the SEL dataset. As mentioned previously, SEL\_Hard images are much more complicated than SEL images. Also, many of SEL images contain only one semantic line. Thus, it is challenging to use only SEL images to learn the harmony between lines in more complicated SEL\_Hard images. Nevertheless, the proposed algorithm yields competitive results to DRM, which performs the best but demands a too high computational cost. Note that the proposed algorithm is about 20 times faster than DRM. Moreover, the proposed algorithm outperforms DRM in terms of AUC\_P.

Figure~\ref{fig:sel_result} compares detection results on the SEL and SEL\_Hard datasets. The conventional algorithms detect redundant lines near object boundaries or  fail to detect implied semantic lines. In contrast, the proposed algorithm detects implied as well as obvious semantic lines more reliably, while ensuring the harmony between detected lines.

\begin{table}[t]\centering

    \renewcommand{\arraystretch}{0.8}
    \caption
    {
        Comparison of the EA-scores (Precision, Recall, F-measure) on the SL5K dataset.
    }
    \vspace*{-0.15cm}
    \resizebox{1\linewidth}{!}{
    \begin{tabular}[t]{+L{2.2cm}^C{1.1cm}^C{1.1cm}^C{1.1cm}^C{1.1cm}}
    \toprule
    & Precision & Recall & F\text{-}measure & HIoU\\
    \midrule
         Zhao \textit{et al.}~\cite{kai2020}             & 70.3 & 74.5 & 72.3 & -\\
         Proposed             & \textbf{79.4} & \textbf{81.4} & \textbf{80.3} & 74.1\\
    \bottomrule
    \\[-15pt]
    \end{tabular}}
    \label{table:sl5k}
\end{table}

\begin{figure}[t]
\vspace{-0.25cm}
\setlength{\belowcaptionskip}{-0.9cm}
    \subfloat {\includegraphics[width=2.05cm,height=1.3cm]{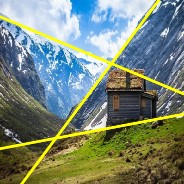}}\,\!\!
    \subfloat {\includegraphics[width=2.05cm,height=1.3cm]{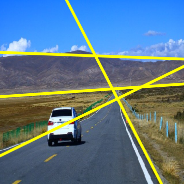}}\,\!\!
    \subfloat {\includegraphics[width=2.05cm,height=1.3cm]{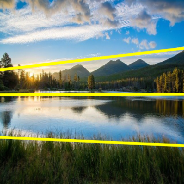}}\,\!\!
    \subfloat {\includegraphics[width=2.05cm,height=1.3cm]{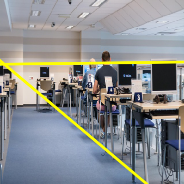}}\\[-4.9ex]

    \subfloat {\includegraphics[width=2.05cm,height=1.3cm]{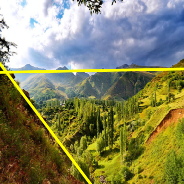}}\,\!\!
    \subfloat {\includegraphics[width=2.05cm,height=1.3cm]{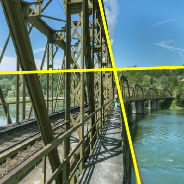}}\,\!\!
    \subfloat {\includegraphics[width=2.05cm,height=1.3cm]{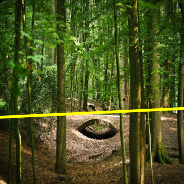}}\,\!\!
    \subfloat {\includegraphics[width=2.05cm,height=1.3cm]{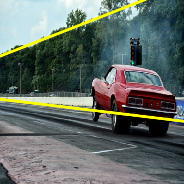}}\\[-4.9ex]

    \setcounter{subfigure}{0}
    \subfloat {\includegraphics[width=2.05cm,height=1.3cm]{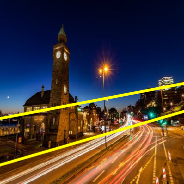}}\,\!\!
    \subfloat {\includegraphics[width=2.05cm,height=1.3cm]{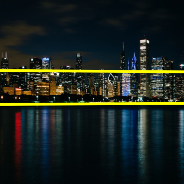}}\,\!\!
    \subfloat {\includegraphics[width=2.05cm,height=1.3cm]{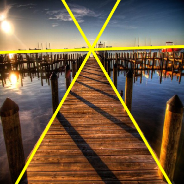}}\,\!\!
    \subfloat {\includegraphics[width=2.05cm,height=1.3cm]{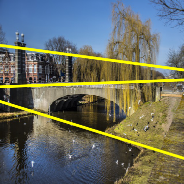}}\\
    \vspace{-0.45cm}
    \caption
    {
        Detection results of the proposed algorithm on the SL5K dataset.
    }
    \label{fig:sl5k_result}
\end{figure}

\vspace*{0.15cm}
\noindent\textbf{Comparison on SL5K:}
Table~\ref{table:sl5k} compares the performances on the SL5K dataset. Zhao \textit{et al.}~\cite{kai2020} report the performances of their algorithm in the EA-score metric only, and their training codes or model parameters are not available. Thus, we compare the results in the EA-score metric only, as done in~\cite{kai2020} . We see that the proposed algorithm outperforms Zhao \textit{et al.}~by significant margins 9.1, 6.9, and 8.0 in terms of precision, recall, and F-measure, respectively. Also, the proposed algorithm yields the HIoU score of 74.1. Figure~\ref{fig:sl5k_result} shows some detection results.

\captionsetup[subfigure]{labelformat=empty}
\begin{figure*}[t]
\vspace{-0.4cm}
\setlength{\belowcaptionskip}{-0.4cm}
    \begin{flushright}

    \subfloat {\raisebox{1.7em}{\rotatebox[origin=t]{90}{\scriptsize Ground-truth}}}\hspace{-0.01cm}\,
    \subfloat {\includegraphics[width=2.42cm,height=1.5cm]{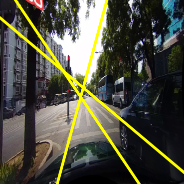}}\,\!\!
    \subfloat {\includegraphics[width=2.42cm,height=1.5cm]{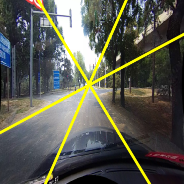}}\,\!\!
    \subfloat {\includegraphics[width=2.42cm,height=1.5cm]{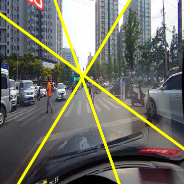}}\,\!\!
    \subfloat {\includegraphics[width=2.42cm,height=1.5cm]{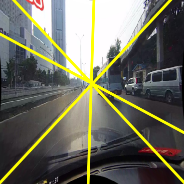}}\,\!\!
    \subfloat {\includegraphics[width=2.42cm,height=1.5cm]{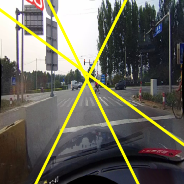}}\,\!\!
    \subfloat {\includegraphics[width=2.42cm,height=1.5cm]{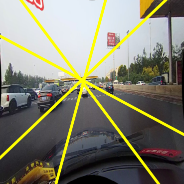}}\,\!\!
    \subfloat {\includegraphics[width=2.42cm,height=1.5cm]{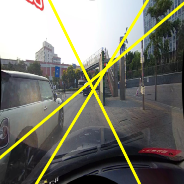}}\\[-2.2ex]

    \subfloat {\raisebox{1.7em}{\rotatebox[origin=t]{90}{\scriptsize Proposed}}}\,\!
    \subfloat {\includegraphics[width=2.42cm,height=1.5cm]{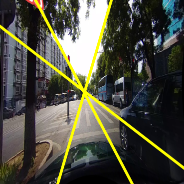}}\,\!\!
    \subfloat {\includegraphics[width=2.42cm,height=1.5cm]{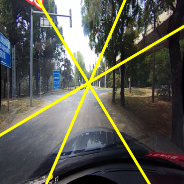}}\,\!\!
    \subfloat {\includegraphics[width=2.42cm,height=1.5cm]{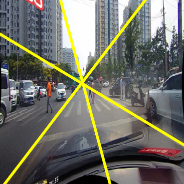}}\,\!\!
    \subfloat {\includegraphics[width=2.42cm,height=1.5cm]{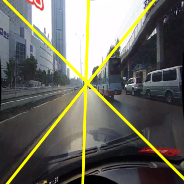}}\,\!\!
    \subfloat {\includegraphics[width=2.42cm,height=1.5cm]{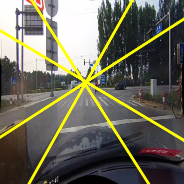}}\,\!\!
    \subfloat {\includegraphics[width=2.42cm,height=1.5cm]{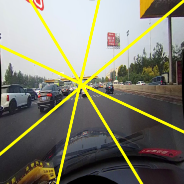}}\,\!\!
    \subfloat {\includegraphics[width=2.42cm,height=1.5cm]{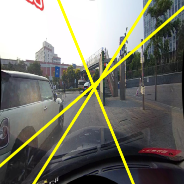}}\\[-2.2ex]

    \setcounter{subfigure}{0}
    \subfloat {\raisebox{1.7em}{\rotatebox[origin=t]{90}{\scriptsize SAD~\cite{hou2019_road}}}}\,\!
    \subfloat {\includegraphics[width=2.42cm,height=1.5cm]{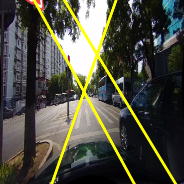}}\,\!\!
    \subfloat {\includegraphics[width=2.42cm,height=1.5cm]{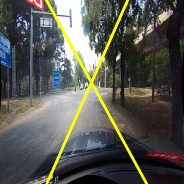}}\,\!\!
    \subfloat {\includegraphics[width=2.42cm,height=1.5cm]{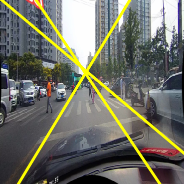}}\,\!\!
    \subfloat {\includegraphics[width=2.42cm,height=1.5cm]{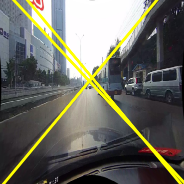}}\,\!\!
    \subfloat {\includegraphics[width=2.42cm,height=1.5cm]{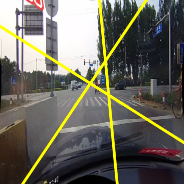}}\,\!\!
    \subfloat {\includegraphics[width=2.42cm,height=1.5cm]{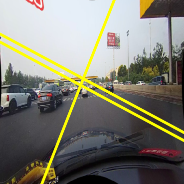}}\,\!\!
    \subfloat {\includegraphics[width=2.42cm,height=1.5cm]{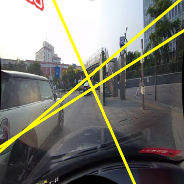}}\\[-2.2ex]

    \subfloat {\raisebox{1.7em}{\rotatebox[origin=t]{90}{\scriptsize UFS~\cite{qin2020}}}}\,\!
    \subfloat {\includegraphics[width=2.42cm,height=1.5cm]{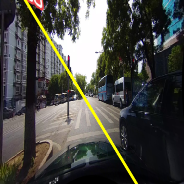}}\,\!\!
    \subfloat {\includegraphics[width=2.42cm,height=1.5cm]{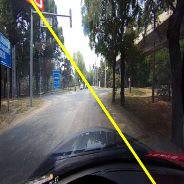}}\,\!\!
    \subfloat {\includegraphics[width=2.42cm,height=1.5cm]{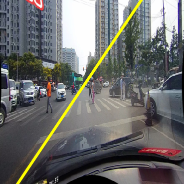}}\,\!\!
    \subfloat {\includegraphics[width=2.42cm,height=1.5cm]{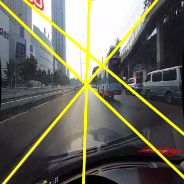}}\,\!\!
    \subfloat {\includegraphics[width=2.42cm,height=1.5cm]{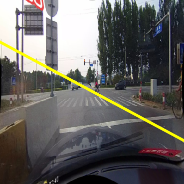}}\,\!\!
    \subfloat {\includegraphics[width=2.42cm,height=1.5cm]{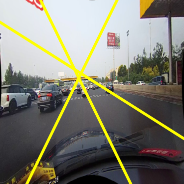}}\,\!\!
    \subfloat {\includegraphics[width=2.42cm,height=1.5cm]{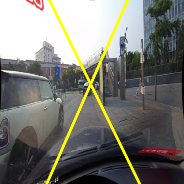}}\\

    \caption
    {
        Comparison of semantic line detection results on the CULane dataset (`no lane' category).
    }
    \label{fig:culane_result}
    \end{flushright}
\end{figure*}
\captionsetup[subfigure]{labelformat=parens}

\vspace*{0.15cm}
\noindent\textbf{Comparison on CULane:}
We compare the proposed algorithm with the conventional road lane detectors~\cite{hou2019_road, qin2020} on the `no lane' category in CULane, in which lanes are implicit or invisible. Conventional techniques are based on the segmentation framework and the ground-truth is also given as a binary mask for each lane. Thus, for comparison, we declare the most overlapping line with the segmentation mask of each lane as a semantic line. The experimental settings are described in detail in the supplemental document. Figure~\ref{fig:culane_result} shows some ground-truth semantic lines and compares their detection results. Although the lines are extremely unobvious, the proposed algorithm detects them more reliably than the conventional detectors. Table~\ref{table:culane} compares the AUC and HIoU scores. Note that, unlike the conventional detectors, the proposed algorithm does not use the information of the maximum number of lanes in a scene. The conventional algorithms poorly recall implied or invisible lanes. The proposed algorithm is slightly less precise, but provides significantly higher recall and F-measure scores than the conventional detectors. Also, the proposed algorithm yields a better HIoU score than the conventional detectors, by exploiting the harmonious property of road lanes, such as parallelness and equal width between adjacent lanes.

\begin{table}[t]\centering

    \renewcommand{\arraystretch}{0.95}
    \caption
    {
        Comparison of the AUC and HIoU scores (\%) on the `no lane' category in the CULane dataset.
    }
    \vspace*{-0.15cm}
    \resizebox{0.92\linewidth}{!}{
    \begin{tabular}[t]{+L{2.2cm}^C{1.1cm}^C{1.1cm}^C{1.1cm}^C{1.1cm}}
    \toprule
    & AUC\_P & AUC\_R & AUC\_F & HIoU\\
    \midrule
         UFS~\cite{qin2020}             & \underline{93.00} & 83.47 & 87.98 & 72.68\\
         SAD~\cite{hou2019_road}        & \bf{93.64} & \underline{84.20} & \underline{88.67} & \underline{74.77}\\
         Proposed             & 92.43 & \bf{91.66} & \bf{92.04} & \bf{76.46}\\
    \bottomrule

    \end{tabular}}
    \label{table:culane}
\end{table}

\begin{table}[t]\centering
    \renewcommand{\arraystretch}{0.95}
    \caption
    {
        Ablation studies for the S-Net, H-Net, and MWCS process on the SEL dataset.
    }
    \vspace*{-0.15cm}
    \resizebox{0.89\linewidth}{!}{
    \begin{tabular}[t]{+C{0.2cm}^L{4.4cm}^C{1.0cm}^C{1.0cm}}
    \toprule
    & & AUC\_F & HIoU\\
    \midrule
         \RomNum{1}. & S-Net                                     & 77.75 & 69.03\\
         \RomNum{2}. & S-Net+H-Net+MWCS\scriptsize (w/o IRC)                   & 84.66 & 79.14\\
         \RomNum{3}. & S-Net+H-Net+MWCS\scriptsize (w/o offset)                   & 86.60 & 80.33\\
         \RomNum{4}. & S-Net+H-Net+MWCS                                 & 86.83 & 81.03\\
    \bottomrule
    \\[-20pt]
    \end{tabular}}
    \label{table:ablation}
\end{table}

\vspace*{0.15cm}
\noindent\textbf{Running time analysis:}
Table~\ref{table:auc_sel} also compares the running times. We use a PC with Intel Core i5-8500 CPU and NVIDIA RTX 2080 ti GPU. Note that SLNet and DRM require a lot of time to extract discriminative line features. Especially, DRM is the slowest method at 1.05 fps, because its mirror attention module and iterative ranking-and-matching process are too demanding. The proposed algorithm and DHT are much faster. Although DHT is the fastest, its recall performance is not competitive.

\subsection{Ablation studies}
We conduct ablation studies to analyze the efficacy of the proposed S-Net, H-Net, and MWCS process on the SEL dataset. Table~\ref{table:ablation} compares several ablated methods. Method \RomNum{1} uses S-Net only to detect semantic lines, in which the selection-and-removal process is performed iteratively until the maximum probability becomes lower than 0.5. Method \RomNum{2} uses H-Net and the MWCS process as well, but H-Net is trained without employing the IRC module. In Method \RomNum{3}, line offsets are not used to refine detection results. Method \RomNum{1} is significantly inferior to the other methods, indicating that both H-Net and MWCS are essential for detecting harmonious semantic lines. Also, by comparing \RomNum{2} with \RomNum{4}, we see that the inter-region correlation feature is effective for estimating the harmony between two lines. Also, from \RomNum{3} with \RomNum{4}, note that the performance is improved by refining detected lines using regression offsets.

\section{Conclusions}
We proposed a novel semantic line detector. First, we developed S-Net to compute the line probabilities and offsets of line candidates. Second, we filtered out irrelevant lines through a selection-and-removal process. Third, we constructed a complete graph, whose edge weights were computed by H-Net. Finally, we determined a maximal weight clique representing a group of harmonious semantic lines. Also, to assess the overall harmony of detected lines, we proposed a novel metric called HIoU. It was experimentally demonstrated that the proposed algorithm can detect harmonious semantic lines effectively and efficiently.

\section*{Acknowledgements}

This work was supported in part by the National Research Foundation of Korea (NRF) through the Korea Government (MSIT) under grant NRF-2018R1A2B3003896 and in part by the 42dot Inc.

{\small
\bibliographystyle{ieee_fullname}
\bibliography{2021_CVPR_DKJIN}
}

\end{document}